\documentclass[runningheads]{llncs}
\usepackage{graphicx}

\usepackage{tikz}
\usepackage{comment}
\usepackage{amsmath,amssymb} 
\usepackage{color}

\usepackage[accsupp]{axessibility}  

\graphicspath{{pdf/}}
\usepackage{booktabs}
\usepackage{amsfonts,amssymb}
\usepackage{makecell}
\usepackage{multirow}
\usepackage{array}
\usepackage{algpseudocode}
\usepackage{caption}
\usepackage{makecell}
\usepackage{multirow}
\usepackage{array}
\usepackage{algpseudocode}
\usepackage[misc]{ifsym}
\usepackage[bottom]{footmisc}

\begin{document}
\pagestyle{headings}
\mainmatter
\def\ECCVSubNumber{100}  

\title{Context-Consistent Semantic Image Editing with Style-Preserved Modulation} 

\titlerunning{Context-Consistent Semantic Image Editing with SPM}
%
\author{Wuyang Luo\inst{1} \and
Su Yang\inst{1}\textsuperscript{\Letter} \and
Hong Wang\inst{1} \and
Bo Long\inst{1} \and
Weishan Zhang\inst{2}}

\authorrunning{W. Luo et al.}
%
\institute{Shanghai Key Laboratory of Intelligent Information Processing, School of Computer Science, Fudan University \and
School of Computer Science and Technology, China University of Petroleum\\
\email{\{wyluo18,suyang\}@fudan.edu.cn}\\
\url{https://github.com/WuyangLuo/SPMPGAN}}


\maketitle
\begin{center}
   \includegraphics[width=11.0cm, trim=15 15 15 15,clip]{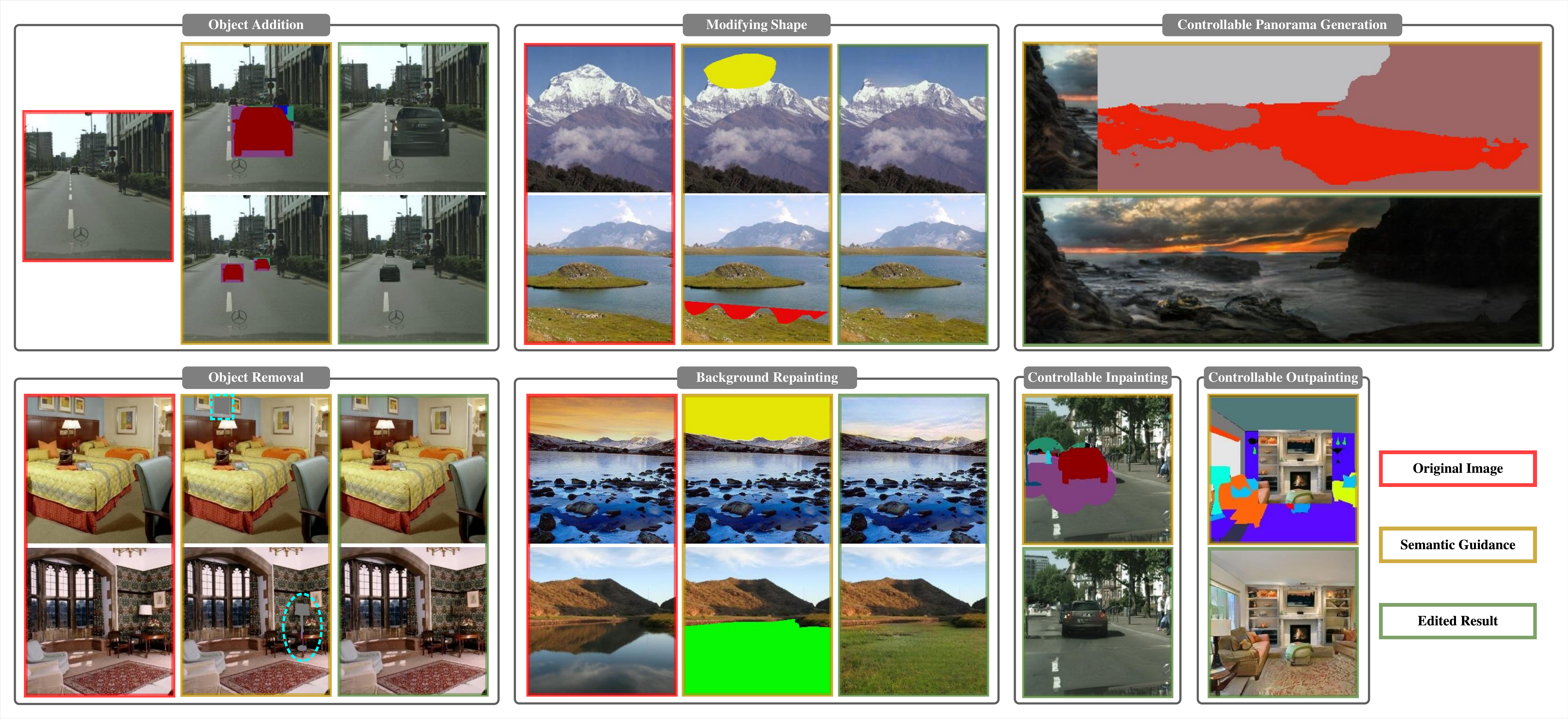}
   \captionof{figure}{Applications of the proposed method. Our image editing system is flexible in responding to a wide variety of editing requirements.}
   \label{fig:app}
\end{center}

\let\thefootnote\relax\footnotetext{\textsuperscript{\Letter} Corresponding author}

\begin{abstract}
Semantic image editing utilizes local semantic label maps to generate the desired content in the edited region. A recent work borrows SPADE block to achieve semantic image editing. However, it cannot produce pleasing results due to style discrepancy between the edited region and surrounding pixels. We attribute this to the fact that SPADE only uses an image-independent local semantic layout but ignores the image-specific styles included in the known pixels. To address this issue, we propose a style-preserved modulation (SPM) comprising two modulations processes: The first modulation incorporates the contextual style and semantic layout, and then generates two fused modulation parameters. The second modulation employs the fused parameters to modulate feature maps. By using such two modulations, SPM can inject the given semantic layout while preserving the image-specific context style. Moreover, we design a progressive architecture for generating the edited content in a coarse-to-fine manner. The proposed method can obtain context-consistent results and significantly alleviate the unpleasant boundary between the generated regions and the known pixels. 
\keywords{Semantic Image Editing, Style-Preserved Modulation}
\end{abstract}

\section{Introduction}
Image editing aims to generate the desired content in a specific region under users' control. This task attracts a lot of research enthusiasm due to its wide application in social media, image and video re-creation, and virtual human-object interaction. The well-known commercial software Photoshop has achieved success in this field. However, the use of such software requires many professional skills and much manual effort.

Most image editing methods fall into a few categories. 
The first category is low-level-guided editing methods \cite{jo2019sc,chen2021deepfaceediting,dong2020fashion,liu2021deflocnet}. They introduce low-level information such as lines and color. These methods can deal with editing simple contours or shapes but only provide very limited editing control and cannot manipulate the high-level semantics of the image. 
The second category is classification-based methods \cite{he2019attgan,hou2022guidedstyle}. They utilize an auxiliary classifier to guide synthesis and edit images. These methods can only control discrete attributes and cannot provide spatial control. 
The third category methods employ GAN inversion technique \cite{tov2021designing,chong2021stylegan,alaluf2021hyperstyle,ling2021editgan}, which relies on a pre-trained GAN and dissects GANs’ latent spaces, finding disentangled latent codes suitable for editing. They require a powerful well-trained StyleGAN, which is impossible in many cases because training a strong StyleGAN \cite{karras2019style,karras2020analyzing} model is not easy, especially for complex scenes. Further, such methods lack flexibility, and the editing of each attribute may require independent training.
The fourth category methods \cite{HIM,SESAME} utilize pixel-level semantic label maps, which define the class labels of pixels in edited regions to control edited content. This task is also known as Semantic Image Editing. Following this line of work, our approach can provide users with greater editing flexibility than the other three categories of methods. Our method includes the following editing capabilities: (1) Our method can be applied to complex scene editing. (2) Users can flexibly edit the image via manipulating semantic layout, such as modifying the shape of objects, adding or removing objects. (3) Edited regions can be selected at arbitrary positions, even beyond the original image boundaries. The Figure \ref{fig:app} demonstrates the versatility of our approach.

\begin{figure}[t]
    \centering
    \includegraphics[width=11.5cm, trim=20 10 10 10,clip]{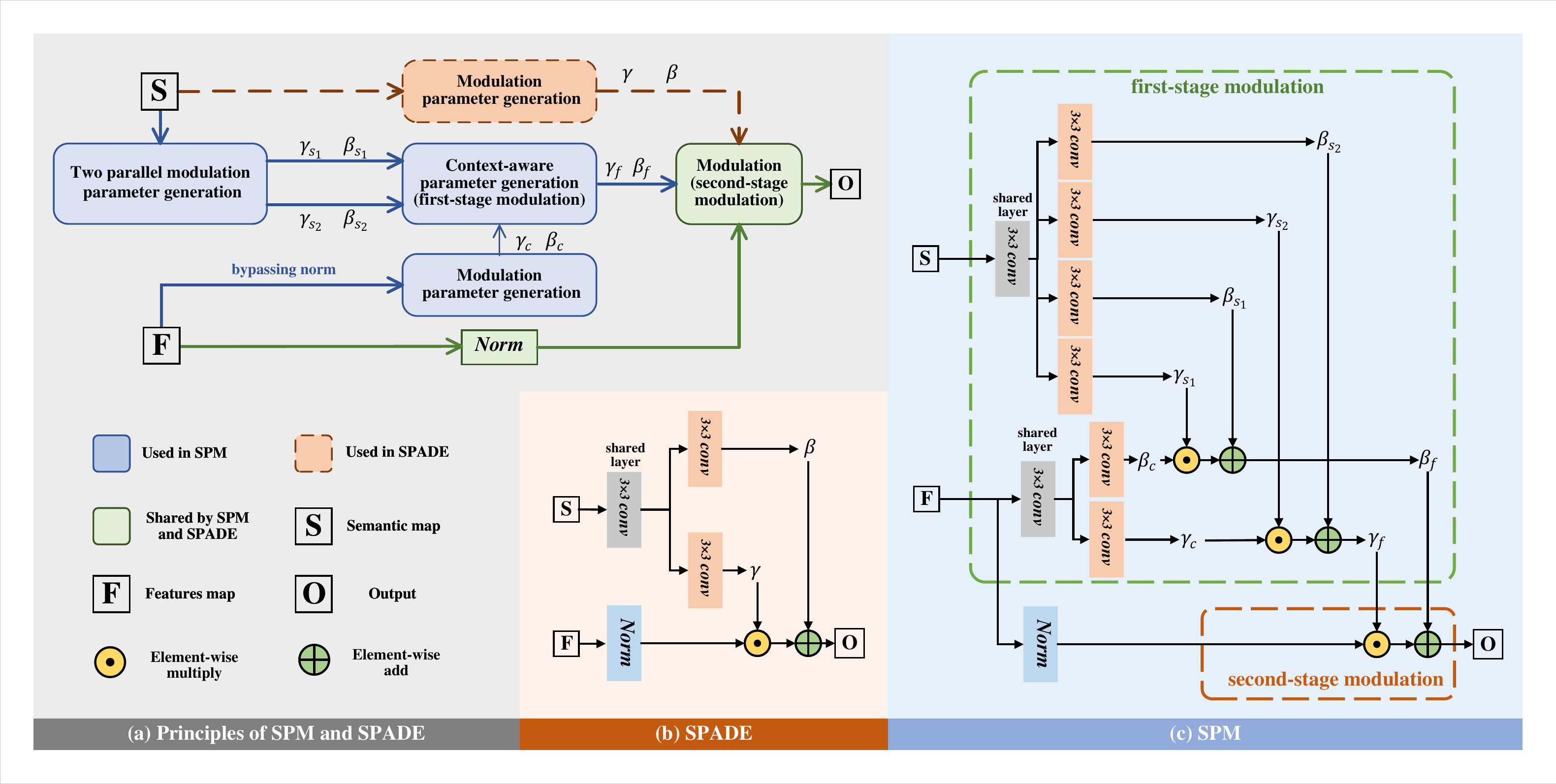}
    \caption{(a) Principle difference between SPM and SPADE; (b) The structure of SPADE; (c) The structure of the proposed SPM.}
    \label{fig:SPADE_SPM}
\end{figure}

Semantic image editing is a non-trivial task. Its challenge lies in keeping context style consistent between edited and known regions. Here, ”context” refers to the non-edited region of the input image, and ”style” is the features
of ”context” such as color/texture. The previous state-of-the-art method SESAME \cite{SESAME} leverages SPADE block \cite{SPADE} to build their generator. SPADE is remarkably effective in conditional image synthesis. Conditional image synthesis learns a mapping from the semantic map domain to the real image domain, synthesizing the entire image according to the given semantic label map. Therefore, the generator may synthesize simple textures to get visually plausible results. However, since known pixels and fake pixels coexist for the image editing task, our task becomes tougher in that the requirement is synthesizing realistic textures and retaining consistency to the context style. Aside from that, image synthesis requires a full semantic label map, but semantic image editing can only see the semantic layout of the edited region. Thus, if SPADE is employed directly on the editing task, only meaningless modulation parameters would be generated in the known region. Previous work \cite{SESAME} often causes significant style inconsistency and unpleasant boundaries for the above reasons. 

To address such limitations of the existing works, we propose a style-preserved modulation module (SPM). Compared with SPADE, which only utilizes one modulation operation, SPM consists of a two-stage modulation process. 
Inspired by the style transfer \cite{adain}, which show that non-normalized feature maps contain high-level ”style” information, we use non-normalized feature maps for context preserving via "bypassing norm".
The principle difference between SPM and SPADE is illustrated in Figure \ref{fig:SPADE_SPM}(a) and their details are described in the section 3. 
Specifically, we first generate two parallel pairs of modulation parameters from semantic maps and a pair of modulation parameters from feature maps. Then we fuse them through the first modulation operation to generate two context-aware modulation parameters. 
The second stage modulation uses the context-aware modulation parameters to modulate feature maps. Through two-stage modulation, SPM can effectively integrate external semantic maps while preserving the image-specific context style.

SPM involves feature maps into the modulation process for preserving contextual style. For image editing tasks, the input is empty in the edited region. The contextual information of the known region is gradually transferred to the edited region through the enlargement of the receptive field of the generator. In order to make the edited region more effectively perceive the contextual style to generate context-aware modulation parameters of SPMs, we build a coarse-to-fine structure to decompose the editing process into multiple scales in a progressive manner. Specifically, we employ multiple generators to receive inputs of different scales. A downsampled version of the input image is fed into the first generator to produce the coarsest result, which contains the coarse-grained image-specific style of the edited region. Subsequent generators can utilize previous results to effectively preserve the contextual style via SPM and refine the detailed textures. 
Our contributions are summarized as follows:
\begin{itemize}
\item[$\bullet$] We propose a context style-preserved modulation for the semantic image editing task, which can inject the layout of the external semantic label map while preserving the image-specific context style. The experiment shows the remarkable effect of SPM for alleviating the inconsistency.
\item[$\bullet$] We build the progressive generative adversarial networks with SPMs for coarse-to-fine generation of edited regions.
\item[$\bullet$] Extensive qualitative and quantitative experiments conducted on several benchmark datasets indicate that our model outperforms the state-of-the-art methods, especially in the sense of contextual style consistency. 
\end{itemize}

\section{Related Works}
\subsection{Image-to-Image Translation}
Image translation attempts to learn a mapping from a source domain to a target domain for realizing the synthesis of new images in the target domain conditioned on source domain images. Image translation can be applied to various tasks, such as image synthesis \cite{zhan2019spatial,zhan2019esir,zhan2018verisimilar}, image editing \cite{HIM,chen2021deepfaceediting,jo2019sc}, style transfer \cite{gatys2016image,adain}, image inpainting \cite{pathak2016context,GC,CA}, image extension \cite{Boundless,yang2019very}, and image super-resolution \cite{lai2017deep,ledig2017photo}. Existing works utilize different conditional inputs as source domain such as semantic label maps, scene layouts, key points, and edge maps. Among them, the most relevant subtask is semantic image synthesis, which aims at generating photo-realistic images conditioned on input semantic label maps.

Semantic image synthesis has achieved remarkable progress benefitting from Generative Adversarial Networks \cite{GAN}. Pix2pix \cite{pix2pix} is the seminal work that introduces a general image translation framework based on conditional generative adversarial networks \cite{cGAN}. The following work pix2pixHD \cite{pix2pixHD} is devoted to generating high-resolution images. SPADE \cite{SPADE} proposes a spatially-adaptive normalization that learns transformation parameters from the semantic layout to modulate the activations in normalization layers. CLADE \cite{CLADE} proposes a lightweight class-adaptive normalization to improve the efficiency of SPADE. Semantic image synthesis has been applied to different downstream tasks in recent works, such as semantic image editing \cite{SESAME}, semantic view synthesis \cite{huang2020semantic}, portrait editing \cite{SEAN,lee2020maskgan}.

\subsection{Semantic Image Editing}
Semantic image editing refers to users provide semantic label maps as an clue to edit the local region of a given image at pixel level. Semantic concepts are more intuitive and fundamental image features than colors, edges, key points, and textures. By manipulating the semantic label map, users can easily edit the image content in many ways, including re-painting, adding, removing, and out-painting semantic information. Unlike semantic image synthesis, which has been extensively studied in recent years, Semantic image editing has not been fully developed because it is challenging. Semantic image editing requires that the edited content not only has high fidelity but also must be consistent with the style of the remaining region. HIM \cite{HIM} is the earliest attempt at this task. HIM first learns to generate the semantic label maps given the object bounding boxes. Then, it learns to generate the edited image from the predicted label maps. HIM can handle object addition and removal. However, HIM can only operate on one foreground target each time. Furthermore, HIM requires a full semantic label map of the entire image as input, which is inconvenient for users. SESAME \cite{SESAME} only input the semantic label map of the edited region, making the image editing tool more practical. In order to improve the quality of the input image, SESAME improves the generator and discriminator. SESAME builds its generator with SPADE and uses a new discriminator to process the semantic and image information in separate streams. Although the previous methods can synthesize plausible results, they ignore the consistency of the context between the edited region and the known region. In contrast, our work is dedicated to reducing this inconsistency.

\subsection{Modulation Technique}
Modulation is also called denormalization, which is an effective way to inject external control information. Unlike the unconditional normalization technique, such as BN \cite{BN}, IN \cite{IN}, and GN \cite{GN}, modulation techniques require external data and follow a similar operating flow. First, feature maps are normalized to zero mean and unit deviation using an unconditional normalization layer. Then the normalized feature maps are modulated with scaling and shifting parameters learned from external data. Modulation techniques were initially applied to style transfer tasks, such as AdaIN \cite{adain} and later adopted in various vision tasks \cite{stylegan,huang2018multimodal,perez2018film}. AdaIN only learned global style representation. To handle external data with spatial dimensions, \cite{SPADE} proposes SPADE for semantic image synthesis. SPADE achieved impressive success in semantic image synthesis. However, the previous methods only consider the external conditional input and ignore the internal contextual information, which is a fatal disadvantage for the semantic image editing task. This paper proposes a new modulation scheme that can aggregate internal context style and external semantic layout. The experimental results show that the proposed method can effectively preserve the context style and improve consistency for semantic image editing.

\section{Approach}

We describe our approach from bottom to top. We first analyze the limitations of SPADE for semantic image editing and introduce SPM proposed in this paper. Then, we introduce how to build a progressive architecture based on SPM.

\subsection{Rethinking SPADE for Semantic Image Editing}
SPADE is a state-of-the-art modulation technology remarkably successful in semantic image synthesis, as shown in Figure \ref{fig:SPADE_SPM}(b). $F^{i} \in \mathbb{R}^{N \times C \times H \times W}$ is the input feature maps of the i-th layers. $N$ is the number of samples in one batch. $C$ is the number of channels. $H$ and $W$ represent the height and width, respectively. SPADE learns two modulation parameters, scaling parameters $\gamma$ and shifting parameters $\beta$, via two convolutional layers from the given semantic label map $S$. First, $F^i$ is normalized in the channel-wise manner:

\begin{equation}
\bar{F^{i}}=\frac{F^{i}-\mu^{i}}{\sigma^{i}}
\end{equation}

\noindent
where ${\mu}^{i} \in \mathbb{R}^{N \times C \times 1 \times 1}$ and ${\sigma}^{i} \in \mathbb{R}^{N \times C \times 1 \times 1}$ are the channel-wise means and standard deviations of $F^{i}$. Then, we perform the modulation operation:
\begin{equation}
\widetilde{F^{i}}= \left(\mathbf{1}+\gamma\right) \odot \bar{F^{i}}+\beta
\end{equation}

Previous work \cite{SESAME} applies SPADE for semantic image editing. However, SPADE is ill-fitted for semantic image editing for the following two reasons: First, SPADE can only generate image-independent modulation parameters from the given external semantic label map. Thus, if two edited images are given the same semantic label map, SPADE will generate the same modulation parameters. This is unreasonable because SPADE ignores image-specific style. Second, for semantic image editing, the generator can only see the semantic layout of the edited region, and the semantic labels of the rest known regions are set to a fixed value. Therefore, SPADE cannot learn effective parameters on the known region. If we naively transfer SPADE to semantic image editing, the above two limitations will cause style inconsistency and unpleasant boundaries.

\subsection{Style-Preserved Modulation}
To solve the issues mentioned above, we propose a two-stage modulation mechanism for style preserving, as shown in Figure \ref{fig:SPADE_SPM}(c). The first stage of modulation aims to integrate the context style and the external semantic layout. The second stage of modulation is to inject the fused information into feature maps.

In the first modulation, we generate two kinds of parameters: Four semantic modulation parameters and two context modulation parameters. Semantic modulation parameters include two groups: $(\gamma_{s_{1}}, \beta_{s_{1}})$ and $(\gamma_{s_{2}}, \beta_{s_{2}})$. The context modulation parameters $(\gamma_{c}, \beta_{c})$ are generated from the original feature maps without passing through the normalization layer. The previous style transfer works \cite{adain} revealed that the style of the image could be washed away by normalization layers. The non-normalized feature maps can retain the context style more. So, we use the original feature maps to generate two context modulation parameters. Finally, 
we perform the first modulation to generate the fused modulation parameters $\gamma_{f}$ and $\beta_{f}$:
\begin{equation}
{\gamma_{f}}= \left(\mathbf{1}+\gamma_{s_{2}}\right) \odot {\gamma_{c}}+\beta_{s_{2}}
\end{equation}
\begin{equation}
{\beta_{f}}= \left(\mathbf{1}+\gamma_{s_{1}}\right) \odot {\beta_{c}}+\beta_{s_{1}}
\end{equation}

\noindent
where $\odot$ denotes element-wise multiplication. All modulation parameters have the same shape as the feature maps $F^i$.

In the second modulation, we use fused modulation parameters to modulate the normalized feature maps $\bar{F^{i}}$.

\begin{equation}
\widetilde{F^{i}}= \left(\mathbf{1}+\gamma_{f}\right) \odot \bar{F^{i}}+\beta_{f}
\end{equation}

Through two-stage modulation process, SPM overcomes the two shortcomings of SPADE: First, the fused modulation parameters integrate the external semantic layout and retain the internal context style. Second, the fused modulation parameters can  generate meaningful modulation parameters for known regions.

\begin{figure}[t]
    \centering
    \includegraphics[width=12.2cm, trim=20 10 10 10,clip]{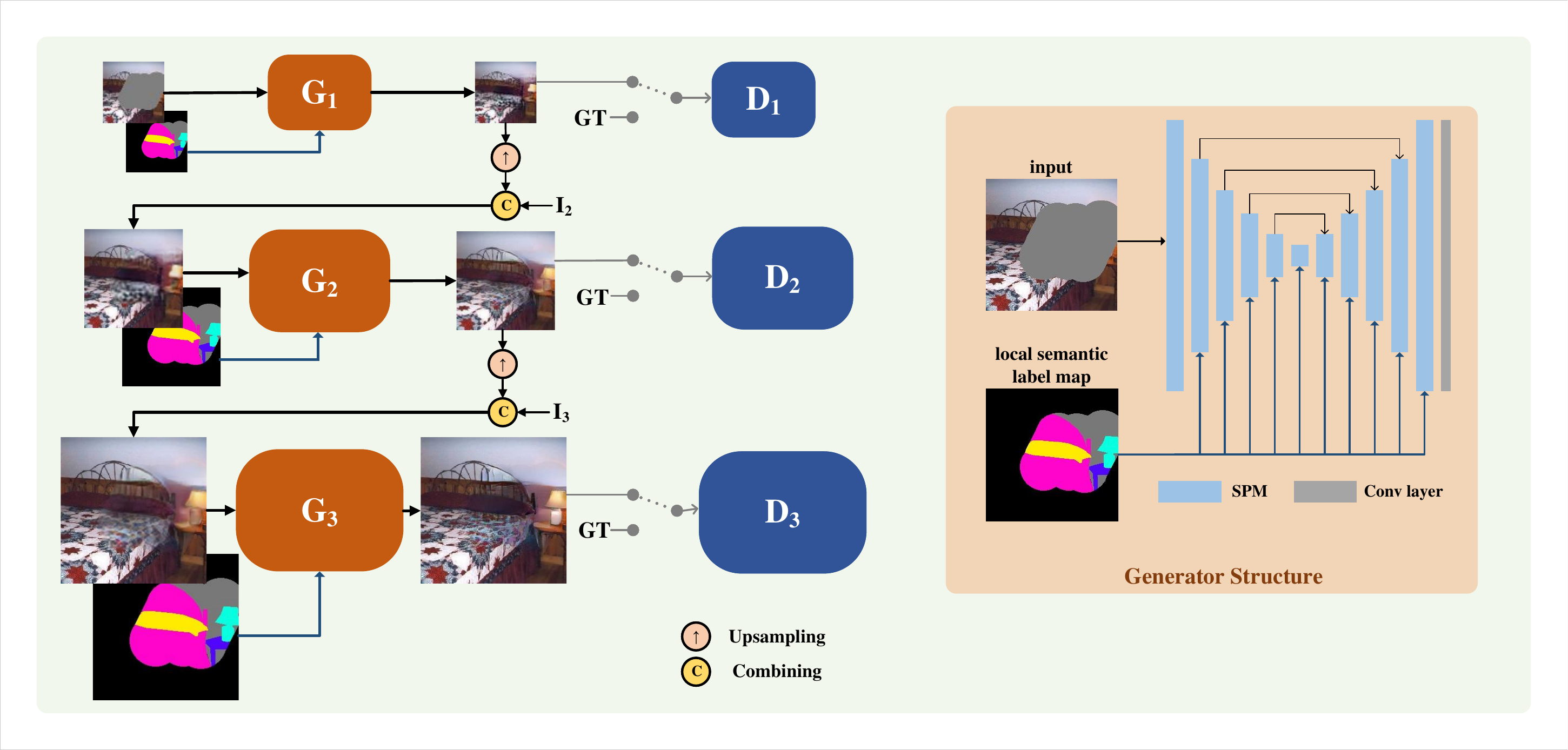}
    \caption{Overview of the progressive architecture.}
    \label{fig:framwork}
\end{figure}

\subsection{Progressive Editing Architecture}
We propose a progressive architecture for image editing based on SPM, called \emph{SPMPGAN}. 
Our model has three inputs: (1) The input image $I \in \mathbb{R}^{256 \times 256 \times 3}$ which contains only known pixels with masked edited region; (2) the local semantic map $S$ providing the semantic layouts of the edited region; and (3) the corresponding mask map $M$ whose value is 0 in the non-edited region and 1 in the edited region. 
Our progressive architecture consists of a pyramid of generators $\left\{G_{1}, G_{2}, G_{3}\right\}$ and discriminators $\left\{D_{1}, D_{2}, D_{3}\right\}$ with an image pyramid of $I$: $\left\{I_{1}, I_{2}, I_{3}\right\}$, where $I_{n}$ is a downsampled version of $I$ by a factor $2^{3-n}$, mask pyramid of $M$: $\left\{M_{1}, M_{2}, M_{3}\right\}$, and semantic map pyramid of $S$: $\left\{S_{1}, S_{2}, S_{3}\right\}$. Each generator $G_{n}$ is trained with an associated discriminator $D_{n}$. $G_{n}$ learns to generate realistic new content in the edited region and try to fool the corresponding discriminator. $D_{n}$ attempts to distinguish the edited result and the real image. We adopt an encoder-decoder architecture with skip connections \cite{unet} for all generators, as shown in Figure \ref{fig:framwork}. Each generator adds a down-sampling layer in the encoder and an up-sampling layer in the decoder on the previous generator. Inspired by\cite{GC}, the discriminators are composed of several convolutional layers with $5 \times 5$ convolution kernel and spectral normalization \cite{SN}. The number of layers of $D_{1}$, $D_{2}$, and $D_{3}$ are 4, 5, and 6, respectively. Thus, each $D_{n}$ has the receptive field with the size of the input $I_{n}$ and captures the entire image's feature. The generation process starts at the coarsest $G_{1}$ and sequentially passes through $G_{2}$ and $G_{3}$ to the original scale. Specifically, the original input $I$ is downsampled to $64 \times 64$ to get $G_{1}$'s input: $I_{G 1}=I_{1}$, and $G_{1}$'s output is $O_{1}$. Then, we combine the upsampled $O_{1}$ with $I_{2}$ as $G_{2}$'s input: $I_{G 2}=O_{1} \odot M_{2}+I_{2}  \odot\left(1-M_{2}\right)$. All generators and discriminators have independent weights. 

\subsection{Training}
We train our progressive model in an end-to-end manner. The training objective for the n-th generator is comprised of a reconstruction loss and an adversarial loss $\mathcal{L}_{\mathrm{adv}}$. The reconstruction loss consists of L1 distance loss $\mathcal{L}_{\mathrm{1}}$ and perceptual loss $\mathcal{L}_{\mathrm{p}}$ \cite{johnson2016perceptual}. We employ the hinge version adversarial loss \cite{brock2018large,liu2019few}. The overall loss can be written as:

\begin{equation}
\mathcal{L} =  \mathcal{L}_{\text {1}} + 10.0\mathcal{L}_{\text {p}} + \mathcal{L}_{adv}
\end{equation}

\section{Experiments}
\subsection{Datasets}
\noindent
\textbf{ADE20K-room}
ADE20K \cite{zhou2017scene} has over 20,000 images together with detailed semantic labels of 150 classes. We select a subset of the ADE20K comprised of  {\tt Bedroom}, {\tt Hotel Room}, and {\tt Living Room}. This subset is called ADE20K-room. We resize all the images with their longer sides no more than 384 and their shorter sides no less than 256. We crop them to $256 \times 256$ when training. This dataset has 2246 images for training and 255 for testing.

\noindent
\textbf{ADE20K-landscape}
We also selected the landscape subclass from ADE20K and use the same preprocessing approach. The difference is that this dataset has only background and no foreground objects. The training set and the testing set contain 1689 images and 155 images, respectively.

\noindent
\textbf{Cityscapes} \cite{cordts2016cityscapes}
The dataset collects streetscapes of 50 German cities, which contains 33 semantic categories. The training and testing set has 2975 and 500 images, respectively, with a resolution of $2048 \times 1024$. We downsample all images to $512 \times 256$ and crop them to $256 \times 256$ patches.

\begin{figure}[h]
    \centering
    \includegraphics[width=12.3cm, trim=25 10 10 10,clip]{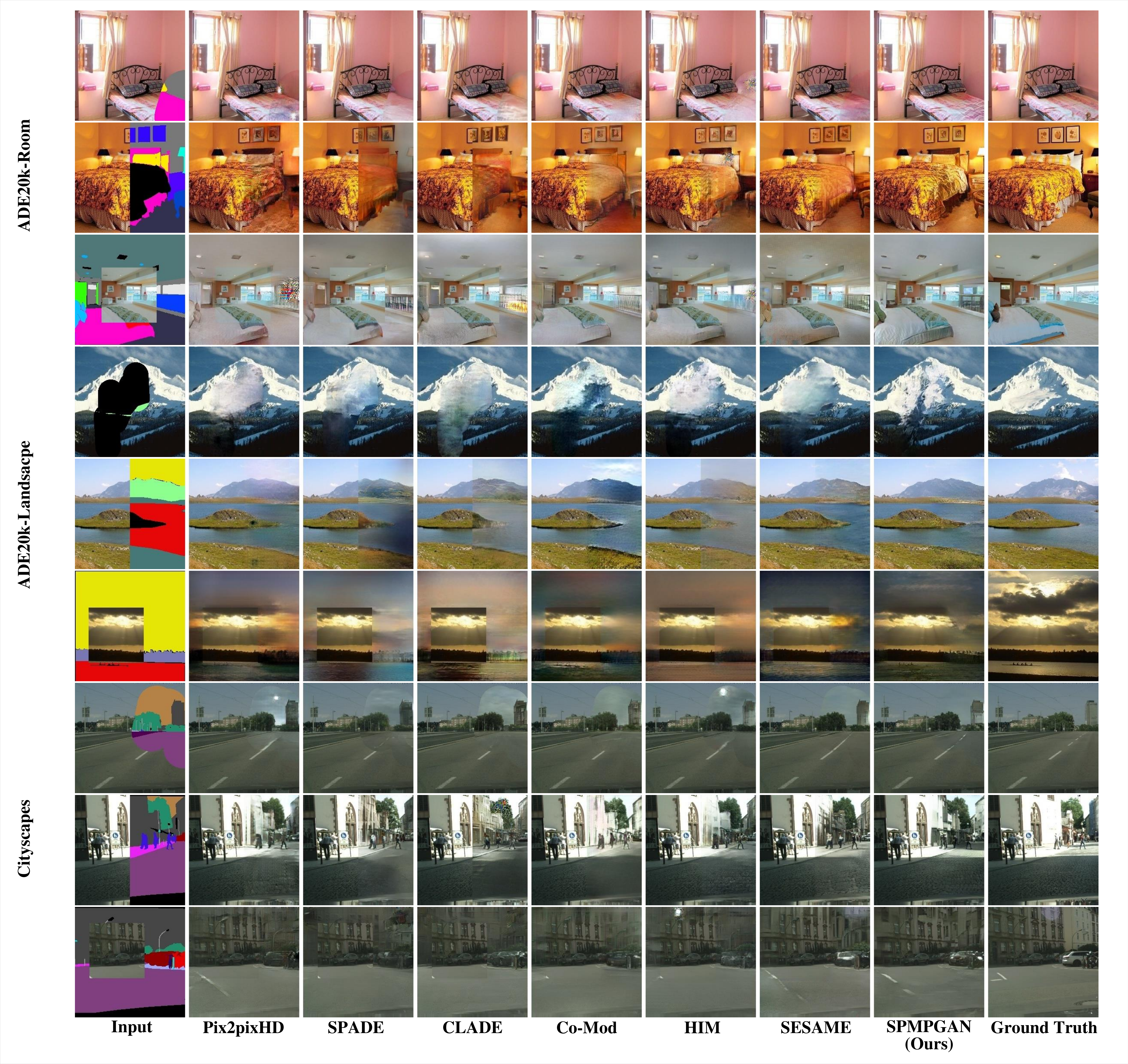}
    \caption{Visual comparison with other methods.}
    \label{fig:sota}
\end{figure}

\begin{figure}[h]
    \centering
    \includegraphics[width=12.3cm, trim=25 10 10 10,clip]{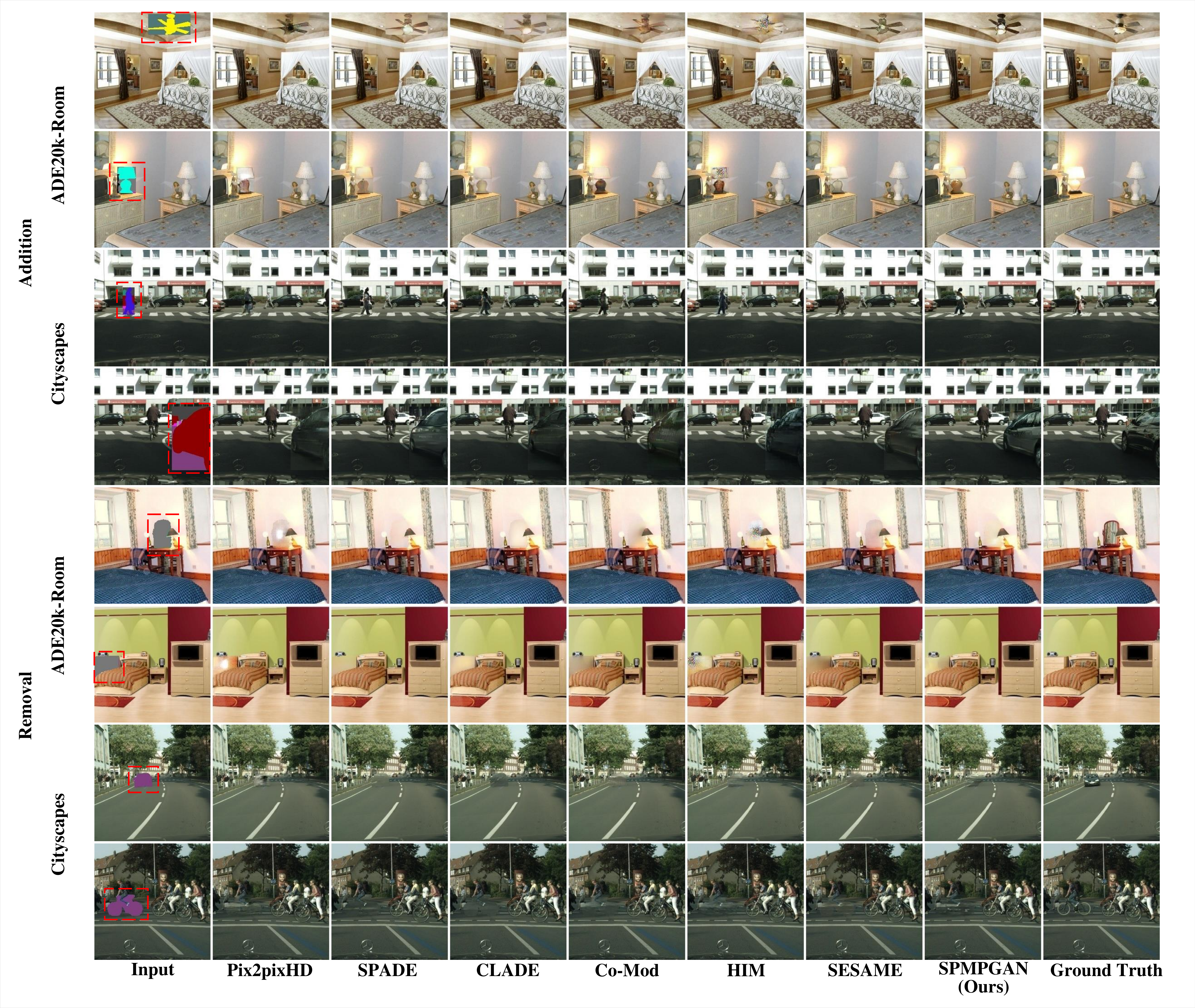}
    \caption{Visual results of addition and removal objects.}
    \label{fig:addition_removal}
\end{figure}

\begin{table}[]
\footnotesize
\centering
\caption{Quantitative comparison with different mask types ($\uparrow$: Higher is better; $\downarrow$: Lower is better).}
\label{tab:sota_ff}
\scriptsize
\setlength{\tabcolsep}{1.0mm}
\tiny
\begin{tabular}{ccccccccccc}
\Xhline{1pt}
\multirow{2}{*}{\begin{tabular}[c]{@{}c@{}}Mask Type\end{tabular}} & \multirow{2}{*}{Method} & \multicolumn{3}{c}{ADE20k-Room}                  & \multicolumn{3}{c}{ADE20k-Landscape}             & \multicolumn{3}{c}{Cityscapes}                   \\ \cline{3-11} 
                                                                     &                         & FID$\downarrow$            & LPIPS$\downarrow$          & mIoU$\uparrow$           & FID$\downarrow$            & LPIPS$\downarrow$          & mIoU$\uparrow$           & FID$\downarrow$            & LPIPS$\downarrow$          & mIoU$\uparrow$           \\
\Xhline{1pt}
\multirow{7}{*}{Free-Form}  & pix2pixHD                                   & 23.72 & 0.107 & 27.49 & 33.90 & 0.120 & 28.30 & 15.28 & 0.090 & 58.69 \\ \cline{2-11}
                            & SPADE                                       & 27.65 & 0.124 & 27.47 & 41.92 & 0.134 & 28.41 & 15.83 & 0.099 & \textbf{59.10} \\ \cline{2-11}
                            & CLADE                                       & 30.77 & 0.126 & 25.91 & 46.59 & 0.139 & 26.39 & 17.06 & 0.103 & 57.72 \\ \cline{2-11}
                            & Co-Mod                                      & 27.37 & 0.111 & 27.52 & 32.35 & 0.124 & 28.60 & 15.88 & 0.097 & 56.50 \\ \cline{2-11}
                            & HIM                                         & 28.64 & 0.133 & 28.04 & 35.89 & 0.116 & 28.43 & 15.58 & 0.093 & 58.99 \\ \cline{2-11}
                            & SESAME                                      & 21.73 & 0.101 & 27.50 & 30.30 & 0.116 & 28.28 & 12.89 & \textbf{0.082} & 58.88 \\ \cline{2-11}
                            & SPMPGAN                                        & \textbf{18.83} & \textbf{0.090} & \textbf{28.22} & \textbf{23.11} & \textbf{0.105} & \textbf{28.73} & \textbf{11.90} & 0.084 & 58.80 \\
\Xhline{0.75pt}

\multirow{7}{*}{Extension}  & pix2pixHD                                   & 38.08 & 0.223 & 27.32 & 56.15 & 0.242 & 28.10 & 26.14 & 0.176 & 58.55 \\ \cline{2-11}
                            & SPADE                                       & 36.43 & 0.211 & 27.62 & 68.96 & 0.277 & 28.44 & 25.78 & 0.194 & 59.01 \\ \cline{2-11}
                            & CLADE                                       & 41.77 & 0.242 & 25.67 & 65.33 & 0.267 & 26.39 & 25.29 & 0.195 & 58.09 \\ \cline{2-11}
                            & Co-Mod                                      & 38.61 & 0.231 & 27.13 & 53.96 & 0.249 & 28.09 & 29.27 & 0.188 & 56.44 \\ \cline{2-11}
                            & HIM                                         & 40.69 & 0.239 & 27.61 & 52.14 & 0.234 & 28.42 & 25.20 & 0.180 & 58.91 \\ \cline{2-11}
                            & SESAME                                      & 36.43 & 0.211 & 27.62 & 48.16 & 0.232 & 28.31 & 20.30 & 0.168 & 59.08 \\ \cline{2-11}
                            & SPMPGAN                                        & \textbf{32.61} & \textbf{0.199} & \textbf{27.73} & \textbf{45.10}  & \textbf{0.217} & \textbf{28.48} & \textbf{19.46} & \textbf{0.167} & \textbf{59.10} \\
\Xhline{0.75pt}

\multirow{7}{*}{Outpainting}    & pix2pixHD                                   & 52.14 & 0.323 & 27.49 & 82.56 & 0.360 & 28.30 & 39.50 & 0.253 & 58.72 \\ \cline{2-11}
                                & SPADE                                       & 47.72 & 0.305 & 27.40 & 88.79 & 0.389 & 28.30 & 33.97 & 0.268 & \textbf{59.07} \\ \cline{2-11}
                                & CLADE                                       & 52.45 & 0.346 & 25.47 & 86.77 & 0.388 & 24.49 & 34.19 & 0.276 & 57.49 \\ \cline{2-11}
                                & Co-Mod                                      & 51.45 & 0.325 & 26.54 & 79.77 & 0.360 & 26.70 & 50.29 & 0.264 & 55.39 \\ \cline{2-11}
                                & HIM                                         & 54.51 & 0.337 & \textbf{28.19} & 77.18 & 0.352 & \textbf{28.57} & 36.27 & 0.252 & 58.99 \\ \cline{2-11}
                                & SESAME                                      & 47.72 & 0.305 & 27.40 & 72.28 & 0.344 & 28.13 & 28.27 & 0.237 & 58.75 \\ \cline{2-11}
                                & SPMPGAN                                       & \textbf{41.52} & \textbf{0.288} & 27.85 & \textbf{63.32} & \textbf{0.328} & 27.56 & \textbf{27.63} & \textbf{0.233} & 58.53 \\
\Xhline{1pt}
\end{tabular}
\end{table}

\begin{figure}[h]
    \centering
    \includegraphics[width=11.5cm, trim=5 10 10 10,clip]{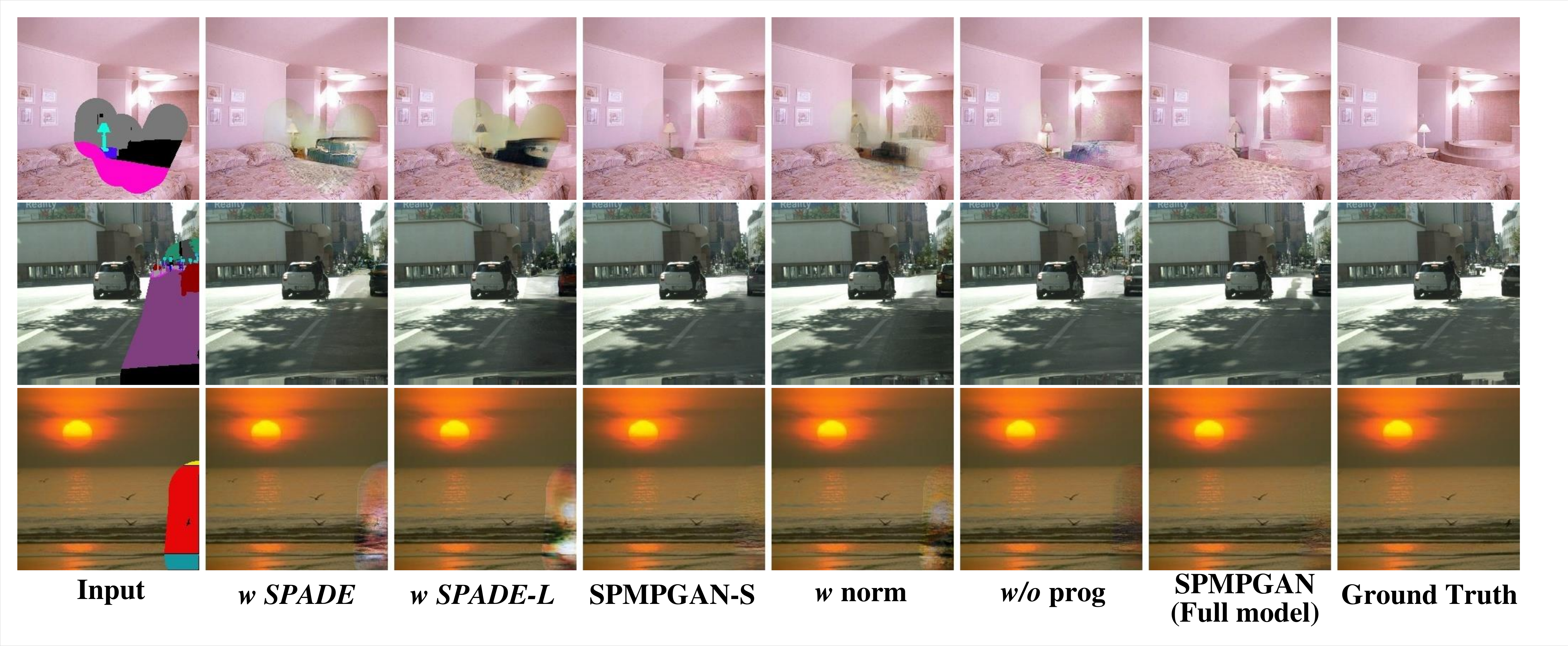}
    \caption{Visual comparison of ablation studies.}
    \label{fig:ablation}
\end{figure}

\subsection{Baselines}
\textbf{Semantic image editing methods}.
We employ two existing works \cite{HIM,SESAME} as baselines. HIM \cite{HIM} introduces a two-stage method for image editing. They first predict semantic layout from object bounding boxes. Then, they generate new content according to the predicted semantic layout. Because in our setting, the ground truth semantic layout of the edited region is known, we directly input the ground truth layout to the second stage of HIM to get the results. SESAME \cite{SESAME} has similar settings with our work. 

\noindent
\textbf{Image synthesis methods}.
Our experiments also include several image generation methods for comparison. These recent works \cite{pix2pixHD,SPADE,CLADE,comod} can be directly transferred to our task via only modifying their generators' input. It is worth mentioning that some recent works cannot be simply adapted for our task. For example, SEAN \cite{SEAN} requires a full segmentation map to calculate their style codes. CoCosNetv2 \cite{zhou2021cocosnet}  requires a full segmentation map to perform their domain alignment. However, our task can only see local semantic label maps. 

\noindent
\subsection{Implementation Details}
To obtain a more flexible model, we employ five types of masks for training: Free-form mask, extension mask, outpainting mask, instance mask, and class mask. The extension mask is the right half of the input. For the outpainting mask, we randomly retain a $128\times128$ patch as the known region. The instance mask contains only a single foreground target, and the class mask drops all the pixels belonging to a semantic class. During training, each mask is randomly selected and sent to the network at each iteration. We use Adam optimizers \cite{adam} for both the generator and the discriminators with momentum $\beta_{1}=0.5$ and $\beta_{2}=0.999$. The learning rates for the generator and the discriminators are set to 0.0001 and 0.0004, respectively. All models are trained for 500 epochs on all datasets. The batch size is set to the maximum value to fit the memory size of a single NVIDIA RTX 3090 GPU.

\subsection{Semantic Image Editing}
We compare our results with state-of-the-art methods using free-form masks, extension masks, and outpainting masks on the three benchmarks. Figure \ref{fig:sota} provides some visual comparisons. Pix2pixHD\cite{pix2pixHD} and HIM\cite{HIM} only use semantic label maps as conditions in the input layer, and they often generate artifacts. SPADE\cite{SPADE}, CLADE\cite{CLADE}, and SESAME\cite{SESAME} can synthesize reasonable structures and realistic textures, but they severely suffer from style inconsistencies leading to unpleasant boundaries. Because they only use the image-independent external semantic map when injecting the semantic label map and completely ignoring the context information. Co-Mod\cite{comod} also has the apparent texture inconsistency as it lacked a specific design for the image editing tasks. The proposed method can effectively integrate the contextual style and the semantic layout to produce realistic textures while preserving the contextual style. Table \ref{tab:sota_ff} also shows the quantitative comparison results. FID \cite{FID} has been widely demonstrated that it is consistent with human visual perception. A lower FID value indicates that results have higher fidelity. LPIPS\cite{LPIPS} evaluates the similarity between the generated image and the corresponding ground truth in a pairwise manner. A lower LPIPS indicates that the generated image is closer to the ground truth. mIoU is employed in the semantic synthesis task \cite{SPADE} to evaluate the alignment between the semantic label map and the generated result. Our method outperforms the other methods in most evaluation metrics.

\subsection{Addition and Removal of Objects}
Our work is capable of adding or removing individual objects by modifying the semantic label maps. Visual results are demonstrated in Figure \ref{fig:addition_removal}. For the object addition, we randomly select an instance of input and extract the boundary boxes to generate its local semantic label map. For object removal, we delete a instance and fill it with nearby background semantic class. Quantitative results shown in Table \ref{tab:addition} indicate that our method achieves the best results in style preservation and fidelity.

\subsection{Controllable Panorama Generation}
A well-trained model can be used recursively to obtain panoramas. Specifically, we employ the generated region of the previous step as the known region of the next step in a sliding window manner. Thus, the input is extended to the right by 128 pixels in each step so that images with arbitrary width can be controllable synthesized. Figure \ref{fig:app} shows a recursive generated result.

\begin{table}[t]
\centering
\caption{Addition and removal results for Cityscapes and ADE20k-Room.}
\label{tab:addition}
\scriptsize
\setlength{\tabcolsep}{1.0mm}
\begin{tabular}{cccccccc}
\Xhline{1pt}
\multirow{2}{*}{Manipulation} & \multirow{2}{*}{Method} & \multicolumn{3}{c}{ADE20k-Room}                 & \multicolumn{3}{c}{Cityscapes}                  \\ \cline{3-8} 
                              &                         & FID$\downarrow$           & LPIPS$\downarrow$          & mIoU$\uparrow$           & FID$\downarrow$           & LPIPS$\downarrow$          & mIoU$\uparrow$           \\ \hline
\multirow{7}{*}{Addition}     & pix2pixHD               & 6.29          & 0.027          & 27.09          & 11.77         & 0.030          & 58.28          \\ \cline{2-8} 
                              & SPADE                   & 5.66          & 0.027          & 27.17          & 10.48         & 0.031          & 58.66          \\ \cline{2-8} 
                              & CLADE                   & 6.21          & 0.028          & 27.16          & 11.03         & 0.031          & 57.55          \\ \cline{2-8} 
                              & Co-Mod                  & 5.75          & 0.026          & 27.23          & 11.28         & 0.031          & 56.40          \\ \cline{2-8}
                              & HIM                     & 9.80          & 0.046          & 27.22          & 11.41         & 0.030          & \textbf{58.75} \\ \cline{2-8} 
                              & SESAME                  & 5.50          & 0.024          & 27.14          & 9.70          & 0.027          & 58.56          \\ \cline{2-8} 
                              & SPMPGAN          & \textbf{5.14} & \textbf{0.022} & \textbf{27.43} & \textbf{9.04} & \textbf{0.026} & 58.68          \\
\Xhline{0.75pt}

\multirow{7}{*}{Removal}    & pix2pixHD               & 4.52     & 0.019     & 28.32    & 15.01    & 0.039    & 55.02    \\ \cline{2-8} 
                            & SPADE                   & 3.96     & 0.019     & 28.35    & 15.48    & 0.040    & 55.04    \\ \cline{2-8} 
                            & CLADE                   & 4.12     & 0.019     & 28.34    & 16.18    & 0.040    & 54.22    \\ \cline{2-8} 
                            & Co-Mod                  & 4.03     & 0.019     & 28.33    & 15.05    & 0.041    & 55.10   \\ \cline{2-8}
                            & HIM                     & 7.44     & 0.035     & 28.33    & 15.10    & 0.040    & \textbf{55.11}    \\ \cline{2-8} 
                            & SESAME                  & 4.02     & 0.018     & 28.34    & 15.52    & 0.041    & 55.08    \\ \cline{2-8} 
                            & SPMPGAN                    & \textbf{3.68}     & \textbf{0.016}     & \textbf{28.35}    & \textbf{14.63}    & \textbf{0.039}    & 55.01    \\ 
\Xhline{1pt}
\end{tabular}
\end{table}

\begin{table}[t]
\centering
\caption{Ablation study with different mask types.}
\label{tab:ablation_ff}
\scriptsize
\begin{tabular}{ccccccccccc}
\Xhline{1pt}
\multirow{2}{*}{\begin{tabular}[c]{@{}c@{}}Mask Type\end{tabular}} & \multirow{2}{*}{Method} & \multicolumn{3}{c}{ADE20k-Room}                  & \multicolumn{3}{c}{ADE20k-Landscape}             & \multicolumn{3}{c}{Cityscapes}                   \\ \cline{3-11} 
                                                                     &                         & FID$\downarrow$            & LPIPS$\downarrow$          & mIoU$\uparrow$           & FID$\downarrow$            & LPIPS$\downarrow$          & mIoU$\uparrow$           & FID$\downarrow$            & LPIPS$\downarrow$          & mIoU$\uparrow$           \\
\Xhline{1pt}
\multirow{6}{*}{Free-Form}   & \textit{w SPADE}         & 23.27     & 0.098     & 27.60     & 34.71     & 0.118     & 28.39     & 14.20     & 0.091     & 58.80          \\ \cline{2-11} 
                             &  \textit{w norm}         & 20.51     & 0.098     & 27.58     & 29.87     & 0.111     & 28.31     & 12.64     & 0.085     & 58.78          \\ \cline{2-11} 
                             & \textit{w/o prog}        & 20.47     & 0.096     & 27.42     & 25.87     & 0.109     & 28.42     & 13.07     & 0.089     & 58.73          \\ \cline{2-11} 
                             & \textit{w SPADE-L}       & 24.11     & 0.098     & 27.61     & 34.68     & 0.116     & 28.43     & 14.40     & 0.090     & 58.79         \\  \cline{2-11}
                             & \textit{SPMPGAN-S}       & 18.93     & 0.090     & \textbf{28.24}     & 23.21     & 0.106     & 28.70     & \textbf{11.89}     & \textbf{0.084}     & \textbf{58.82}         \\ \cline{2-11}
                             & \textit{SPMPGAN}         & \textbf{18.83}     & \textbf{0.090}     & 28.22     & \textbf{23.11}      & \textbf{0.105}     & \textbf{28.73}     & 11.90     & 0.084     & 58.80 \\

\Xhline{0.75pt}

\multirow{6}{*}{Extension}      & \textit{w SPADE}      & 36.84     & 0.220     & 27.51     & 53.02     & 0.239     & 28.92     & 21.99     & 0.173     & 58.88             \\ \cline{2-11}
                                & \textit{w norm}       & 32.76     & 0.205     & 27.56     & 48.43     & 0.228     & 28.92     & 20.50     & 0.176     & 59.01             \\ \cline{2-11}
                                & \textit{w/o prog}     & 33.87     & 0.205     & 27.44     & 45.96     & 0.222     & 28.86     & 21.00     & 0.170     & 59.09             \\ \cline{2-11}
                                & \textit{w SPADE-L}    & 36.14     & 0.218     & 27.48     & 53.13     & 0.240     & \textbf{28.93}     & 21.86     & 0.174     & 58.81             \\ \cline{2-11}
                                & \textit{SPMPGAN-S}    & \textbf{31.92}     & 0.200     & \textbf{27.74}     & 45.17     & 0.218     & 28.47     & \textbf{19.12}     & \textbf{0.167}     & \textbf{59.12}             \\ \cline{2-11}
                                & \textit{SPMPGAN}      & 32.61     & \textbf{0.199}     & 27.73     & \textbf{45.10}     & \textbf{0.217}     & 28.48     & 19.46     & 0.167     & 59.10             \\ \cline{2-11}

\Xhline{0.75pt}

\multirow{6}{*}{Outpainting}    & \textit{w SPADE}      & 47.37     & 0.321     & 28.38     & 71.52     & 0.357     & \textbf{28.84}     & 31.33     & 0.244     & 58.95             \\ \cline{2-11}
                                & \textit{w norm}       & 42.31     & 0.300     & \textbf{28.52}     & 66.52     & 0.337     & 28.82     & 27.74     & 0.235     & 57.98             \\ \cline{2-11}
                                & \textit{w/o prog}     & 43.98     & 0.297     & 28.05     & 66.32     & 0.329     & 27.39     & 29.54     & 0.238     & 58.53             \\ \cline{2-11}
                                & \textit{w SPADE-L}    & 47.16     & 0.318     & 28.39     & 70.33     & 0.354     & 28.83     & 31.43     & 0.243     & \textbf{58.95}              \\ \cline{2-11}
                                & \textit{SPMPGAN-S}    & \textbf{41.49}     & 0.289     & 27.80     & \textbf{62.43}     & 0.330     & 27.63     & \textbf{27.39}     & \textbf{0.228}     & 58.59             \\ \cline{2-11}
                                & \textit{SPMPGAN}      & 41.52     & \textbf{0.288}     & 27.85     & 63.32     & \textbf{0.328}     & 27.56     & 27.63     & 0.233     & 58.53             \\ \cline{2-11}
\Xhline{1pt}

\end{tabular}
\end{table}
\subsection{Ablation Study}
\noindent
\textbf{Style-preserved modulation}

We study the importance of SPM for style preserving. We replace all SPMs with SPADE blocks ("\textit{w SPADE}"). The visual results are shown in Figure \ref{fig:ablation}. It can be observed that SPADE leads to unpleasant boundaries. This is because SPADE completely ignores the image-specific context style and only uses local semantic label maps to modulate feature maps. As a comparison, SPM can relieve the inconsistency. The two-stage modulation can integrate the context style and the external semantic label map. In addition, SPM can also help the generator to synthesize more realistic texture details. We also study the influence of "bypassing norm" for style preserving. Specifically, for the generation of $\gamma_{c}$ and $\beta_{c}$ in SPM, we replace original feature maps by normalized feature maps ("\textit{w norm}"). The experimental results show that the style preserving is significantly weakened. It proves that the normalization operation washes away context style. Therefore, we use the original feature maps without normalization in SPM. Quantitative results are also demonstrated in Table \ref{tab:ablation_ff}.

\begin{table}[t]
\centering
\caption{Users study results.}
\label{tab:user}
\setlength{\tabcolsep}{1.0mm}
\begin{tabular}{lccc}
\Xhline{1pt}
Method     & HIM & SESAME & Ours \\ \hline
preference & 128  & 493     & \textbf{1479}  \\ \bottomrule
\end{tabular}
\end{table}

\noindent
\textbf{Effectiveness of progressive architecture}

\begin{figure}[t]
    \centering
    \includegraphics[width=12.2cm, trim=25 10 10 10,clip]{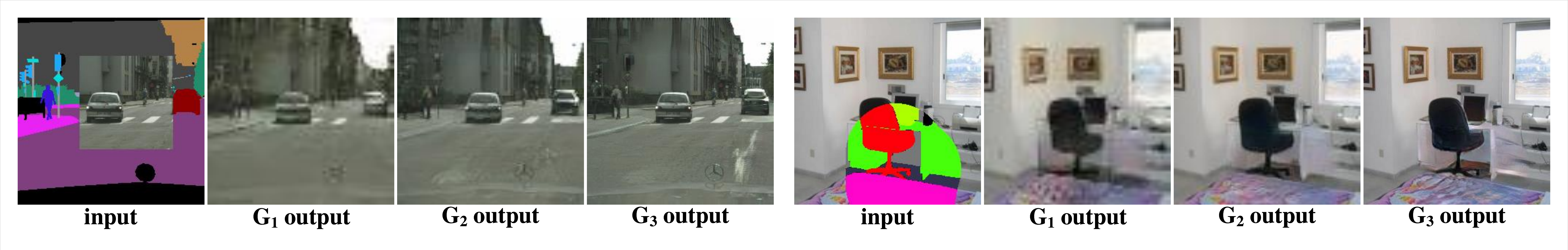}
    \caption{Outputs of all generators.}
    \label{fig:progressive}
\end{figure}

We conduct an ablation study to demonstrate the effectiveness of the progressive design for synthesizing high-quality results. We only use the last level generator as the baseline ("\textit{w/o prog}"). Figure \ref{fig:ablation}(c) shows that without the progressive generation, the model will produce style inconsistency and unrealistic textures. The outputs of the generators of all scales are shown in the Figure  \ref{fig:progressive}. It can be seen, $G_{1}$ synthesizes the global structure, and $G_{2}$ and $G_{3}$ produce the sharper detail. Quantitative results are given in Table \ref{tab:ablation_ff}, which indicates that progressive architecture contributes to performance improvement.

\subsection{User Study}
We invited twenty-one volunteers with image generation expertise to perform the user study. For each volunteer, we randomly selected 100 results of HIM \cite{HIM}, SESAME \cite{SESAME}, and the proposed model from the testing sets with different masks. We asked them to choose an image that better preserves the context style in the edited region. As shown in Table \ref{tab:user}, the results of the proposed method are clearly preferred by the users over the previous methods.

\subsection{Study of Model Scale}
This study demonstrates that our performance improvement stems from the novel design of SPM rather than increasing parameters. As shown in Table \ref{tab:param}, our model follows SPADE to set the number of output channels $C^{h}$ of the shared layer to 128. We reduce $C^{h}$ of all SPMs to 64 and keep the structure unchanged ("\emph{SPMGAN-S}"). We do not observe the performance drop. In addition, we insert more SPADE blocks into "\emph{w SPADE}" to obtain a new baseline "\emph{w SPADE-L}" . The experimental results are shown in the Table \ref{tab:ablation_ff}, "\emph{w SPADE-L}" does not obtain performance gain by simply increasing the network scale and computational consumption. The performance of "\emph{SPMGAN-S}" still significantly outperforms "\emph{w SPADE-L}" with fewer parameters.

\begin{table}[t]
\centering
\caption{Comparison of the number of parameters.}
\label{tab:param}
\begin{tabular}{lcccc}
\Xhline{1pt}
             & \emph{  w SPADE  } & \emph{  w SPADE-L  } & \emph{  SPMPGAN  } & \emph{  SPMPGAN-S  } \\ \hline
ADE20k-Room & 63.4 M  & 90.0 M    & 118.4 M & 76.9 M    \\ \hline
Cityscapes  & 57.8 M  & 81.5 M    & 112.7 M & 74.0 M   \\
\Xhline{1pt}
\end{tabular}
\end{table}

\section{Conclusion}
This paper is dedicated to solving style inconsistency for the semantic editing task. We propose a style-preserved modulation and a progressive architecture that effectively injects the structure from semantic label maps while preserving the context style. The key of SPM lies in effectively integrating contextual information and semantic label maps. We also demonstrate the ability of our method for various applications.

\noindent
\textbf{Acknowledgement} This work is supported by State Grid Corporation of China (Grant No. 5500-202011091A-0-0-00).

\clearpage
%
%
\bibliographystyle{splncs04}
\bibliography{egbib}
\end{document}